\documentclass[runningheads]{llncs}
\usepackage[T1]{fontenc}
\usepackage[pdftex]{graphicx}
\usepackage[font=small,labelfont=bf,tableposition=top]{caption}
\usepackage{subcaption}

\usepackage{booktabs}
\usepackage{multirow}

\usepackage{xspace}

\makeatletter
\DeclareRobustCommand\onedot{\futurelet\@let@token\@onedot}
\def\@onedot{\ifx\@let@token.\else.\null\fi\xspace}

\def\eg{\emph{e.g}\onedot} 
\def\ie{\emph{i.e}\onedot}

\makeatother

\newcommand{\ours}{\mbox{\textsc{SF-MASK}}\xspace}

\usepackage{hyperref}

\begin{document}
\title{A Masked Face Classification Benchmark on Low-Resolution Surveillance Images}
\titlerunning{A Masked Face Classification Benchmark}
%

\author{Federico Cunico\orcidID{0000-0001-9619-9656}
\and
Andrea Toaiari\orcidID{0000-0002-3759-8865}
\and
Marco Cristani\orcidID{0000-0002-0523-6042}
}
\authorrunning{F. Cunico et al.}
%
\institute{Department of Computer Science, University of Verona, Italy \email{[name].[surname]@univr.it}}
\maketitle

\begin{abstract}
We propose a novel image dataset focused on tiny faces wearing face masks for mask classification purposes, dubbed \emph{Small Face MASK} (\ours), composed of a collection made from 20k low-resolution images exported from diverse and heterogeneous datasets, ranging from 7~x~7 to 64~x~64 pixel resolution. An accurate visualization of this collection, through counting grids, made it possible to highlight gaps in the variety of poses assumed by the heads of the pedestrians. In particular, faces filmed by very high cameras, in which the facial features appear strongly skewed, are absent. To address this structural deficiency, we produced a set of synthetic images which resulted in a satisfactory covering of the intra-class variance. Furthermore, a small subsample of 1701 images contains badly worn face masks, opening to multi-class classification challenges. Experiments on \ours focus on face mask classification using several classifiers.
Results show that the richness of \ours (real + synthetic images) leads all of the tested classifiers to perform better than exploiting comparative face mask datasets, on a fixed 1077 images testing set. Dataset and evaluation code are publicly available here: \href{https://github.com/HumaticsLAB/sf-mask}{https://github.com/HumaticsLAB/sf-mask}

\end{abstract}

\section{Introduction}

The COVID-19 epidemic has posed great challenges to the world. The World Health Organization has provided guidelines on tools to use and protocols to adopt to fight this disease \cite{world2021covid}. Among these, distancing and masks have proven to be effective in preventing the spread of the epidemic. In particular, face masks have become the bulwark of COVID-19 prevention,  trapping the droplets exhaled from infected individuals~\cite{forouzandeh2021face}. Masks are crucial especially when distancing cannot be easily held, \eg in small indoor environments or on public transportation~\cite{dzisi2020adherence}. The numbers are impressive: 52 billion disposable face masks were produced in 2020, with a market size which is expected to reach USD 2.1 billion by 2030~\cite{FaceMask2022}. It follows that face masks have become a cultural norm~\cite{malik2021impact}, and will probably be part of the ``new normal'' after the pandemic will definitely end. Actually, coronavirus is expected to stay with us in the future, in the form of an endemic disease - meaning that it will continue to circulate in pockets of the global population for years to come~\cite{phillips2021coronavirus}.

\begin{figure}[h!]
    \centering
    \includegraphics[width=.9\linewidth]{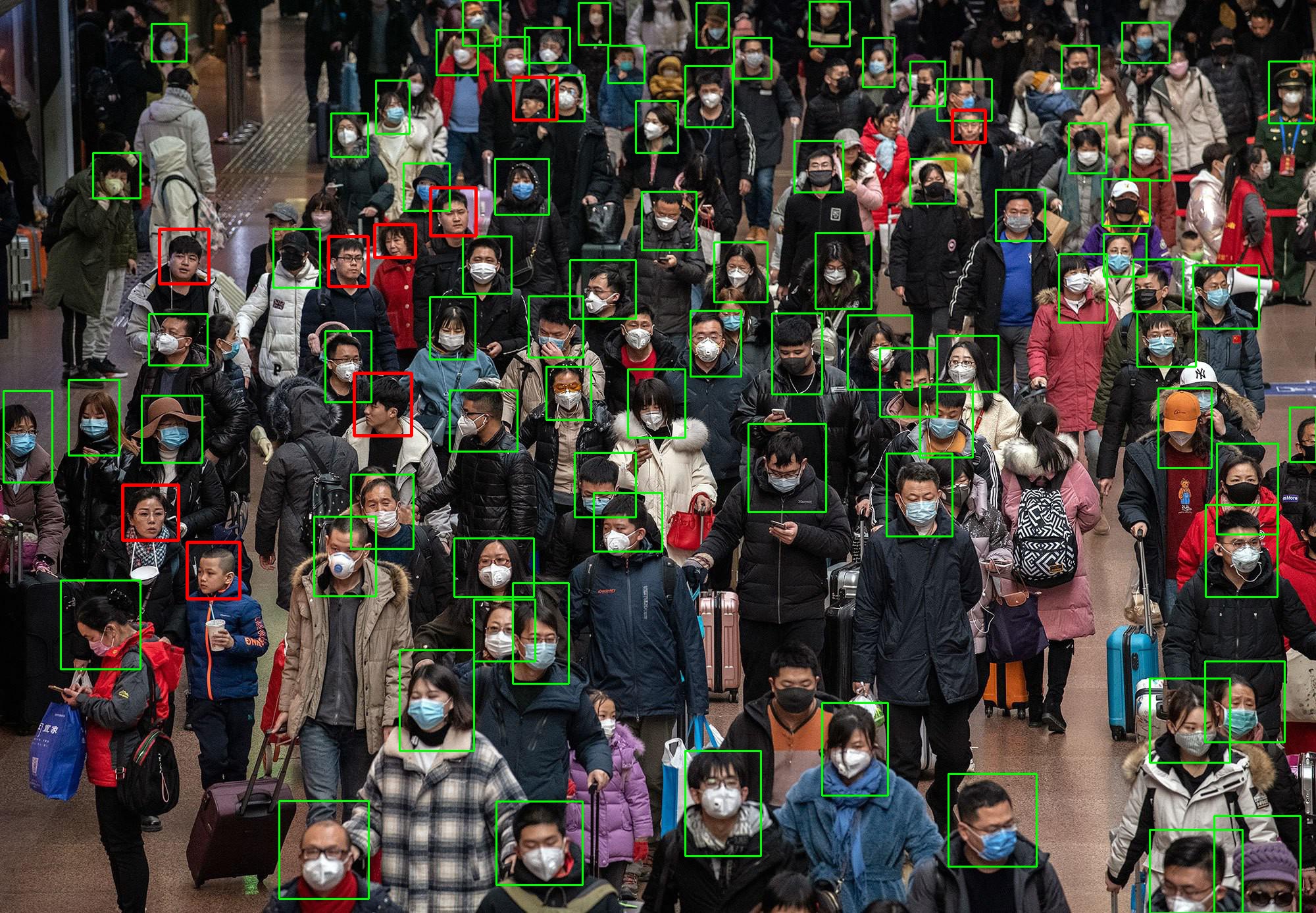}
    \caption{Example of small bounding box classification using a \mbox{ResNet-50} trained on \ours with synthetic augmentation.}
    \label{fig:example}
\end{figure}

\begin{figure*}[h!]
    \centering
    \includegraphics[width=\textwidth]{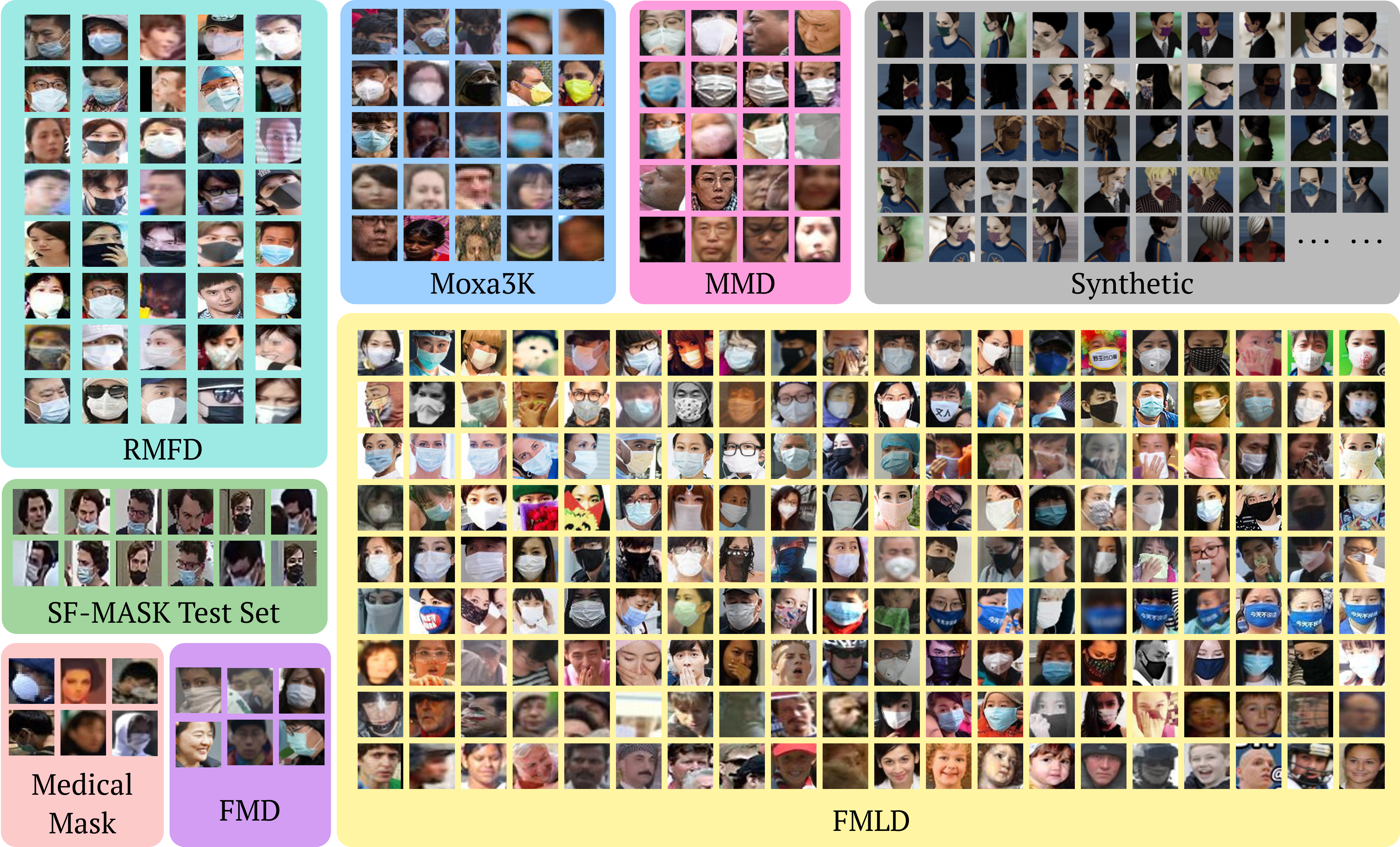}\label{subfig:ours}
    \caption{Composition of \ours dataset, a novel low-resolution masked face dataset. Each box presents a sample of the data extracted from the datasets involved, with the size of the boxes proportional to the portion used.}
    \label{fig:dataset_composition}
\end{figure*}

Face masks are effective in the measure with which they are worn by everyone and in a consistent way~\cite{howard2021evidence,mitze2020face,esposito2020universal,feng2020rational}: for this reason, non-collaborative monitoring to ensure proper use of the face mask seems to be a very effective solution, resulting in an obvious clash between privacy and safety~\cite{pooja2021face,shorfuzzaman2021towards,ram2020mass}. We are not discussing this issue here, focusing instead on the challenge of detecting face masks from a computer vision point of view. Actually, if the privacy and ethical issues were resolved, there would still remain a fundamental problem for video surveillance, namely that automatically identifying face masks is difficult, especially in selected scenarios.  

Fig.~\ref{fig:example} shows an example of this challenge: a crowd of people is captured by a single picture, where each face can be enclosed by a bounding box of few pixels; furthermore, faces can be rotated, subject to occlusion, affected by glares/dark areas. The face masks themselves bring in a new challenge, since they can be of different shapes and colors, patterned or even skin coloured. Finally, there is an underlying chicken-egg problem: capturing the face to check the presence of a mask is made difficult by the mask itself, which covers most of the discriminative features used to detect a face (mouth, nose). For this reason, general approaches for checking the presence of the mask are based on a people detector run in advance, where the face is localized thanks to the body layout prior~\cite{zhang2017detecting}. We assume this pipeline, and specifically to start with an image crop of a face, with the aim of classifying if it wears a mask or not. 

Under this scenario, in this paper we present a novel dataset, dubbed \emph{Small Face MASK} (\ours), which serves to train face mask classification systems, supposed to work in a surveillance scenario, with faces captured at a native low resolution. The dataset is composed of 21,675 real-world RGB images with 9,055 images with face masks correctly worn and 12,620 images without face masks. The data collection has span over all the possible datasets containing small faces with masks in the literature, including \cite{rwfmd,mmd,kagglefmd,kagglemmd,fmld,moxa}, with proportions visualized in Fig.~\ref{fig:dataset_composition}. One of the main contributions of this paper has been that of first understanding \emph{what is contained} in these datasets in terms of intra-class variability (variation in poses, mask colors, etc.). Since an automatic quantitative analysis is hard (for example, head pose estimation to calculate the variation in pose is itself a tough challenge~\cite{jan2018wnet}), we performed a qualitative analysis with the help of the counting grids~\cite{perina2011image} (CG), a manifold learning approach specifically suited for the image data. The CG allowed us to discover the absence of specific poses: actually,  faces filmed by very high cameras, in which the
facial features appear strongly skewed, are absent.  To address
this structural deficiency, we produced a set of 15,481 synthetic images exploiting a pipeline based on Makehuman~\cite{makehuman} + Blender~\cite{blender}, where the masked faces are captured in specific poses. This resulted in the satisfactory completion of the intra-class variance, as showed by dimensionality reduction experiments. Additionally, there are 1,701 faces where masks have been incorrectly worn, meaning that the face mask is visible, but is below the nose or the chin. 

The dataset is divided into a training and a testing partition. The testing partition has been built from scratch to encapsulate all of the issues discussed above (hard illumination, rotated heads, occlusions) in a real multi-camera surveillance environment, offering a valid benchmark for the face mask classification problem.

\ours has been employed to train with different classifiers a face mask classification system.
The results show that the classifiers which have been trained with \ours provide the best performance with respect to alternatives originated by a subset of the whole data. 

Summarizing, the contributions of this paper are:
\begin{itemize}
    \item We collect most of the publicly available datasets on face masks available in the literature, offering a sound qualitative overview of what is contained therein, thanks to the counting grids; 
    \item We crafted \ours, a novel dataset for low-resolution face mask classification, filling the gap of the missing head poses in the available datasets by synthetic data;
    \item We perform exhaustive classification experiments on \mbox{\ours}, showing that it works as an effective training set for the masked face classification task.
\end{itemize}

\section{Related works}\label{sec:related}

In this work, we define \emph{face mask classification} as the task of individuating the presence of a face mask, or its absence, on a test image that tightly encloses a face. With \emph{face mask detection} we indicate a system that, given an image, provides as output the bounding boxes of faces, where each bounding box is associated with a label indicating the presence or absence of a face mask.  In this literature review, we consider datasets used for both applications. On the contrary, we do not consider all those references which deal with face mask recognition, that individuates a person's identity when they are covered by a face mask. Readers interested in face mask recognition may refer to the survey of~\cite{recognition_survey}.

Since the start of the COVID-19 pandemic two years ago, many research contributions have been published in relation to face mask detection and classification~\cite{survey}. 

Most of the approaches are based on CNN networks, which require to have large amounts of data available~\cite{mafa,urban_env,superres,singh_face_2021}.

One of the first public datasets is MAFA~\cite{mafa}, explicitly focused on face detection in the presence of occlusions, more or less pronounced. It contains around 35,000 annotated faces collected from Internet, where occlusion elements are of various nature, and not only face masks, such as human body parts, scarfs, hats or other accessories. An extension of MAFA is MAFA-FMD~\cite{retinafacemask}, which adds a new wearing state for face masks images, \ie \emph{incorrect}, and increases the number of low-resolution images. Another example of dataset including improperly worn masks is~\cite{mmd}, despite being considerably smaller.
In ~\cite{rwfmd}, the authors proposed three different datasets:  MFDD, RMFRD, SMFRD. The former is indicated for the face mask classification and detection tasks: it contains 24,771 masked face images, derived mainly from the web. RMFRD was generated by searching the web for images of celebrities with and without a mask, and it is specifically suited for the face recognition task. Lastly, SMFRD is a simulated dataset, where real face images are enriched with mask patches. 

It is also possible to find small but interesting face mask datasets on the platform Kaggle, such as~\cite{kagglemmd} and~\cite{kagglefmd}.
Moxa3K~\cite{moxa} is a fairly small dataset, consisting of data from~\cite{kagglemmd} and other images found online. In this case, most of the faces are relatively small, since the images contain crowds.

Some common practices to create new and more diverse datasets consist in combining pre-existing masked face collections into one consistent set or combining masked face images crawled from the web with popular face detection and face recognition datasets, such as WiderFace~\cite{widerface} and FFHQ~\cite{ffhq}, which contain many images of faces with various resolutions, lighting conditions and head poses. Some examples of datasets constructed from the combination of previously released collections are LMFD~\cite{fmld},~\cite{singh_face_2021},~\cite{aizoo} (MAFA~\cite{mafa}~+~WiderFace~\cite{widerface}) and~\cite{dey_face} (MFDD~\cite{rwfmd} and SMFRD~\cite{rwfmd}).

In comparison to these datasets, which report faces at various resolution, we wanted to focus on images of natively low-resolution faces. This is to be compliant with video surveillance scenarios, assuming that the final purpose is to check the crowd at a distance for the presence of masks.

\subsection{Synthetic Face Masks Datasets}
The idea of crafting synthetical data as proxy for real ones for a better training set is not novel. 
Some examples of ``augmented'' datasets are SMFRD~\cite{rwfmd}, which uses custom software to apply mask patches to face images from the LFW~\cite{lfw} and Webface~\cite{webface} datasets.
MFNID~\cite{maskedfacenet} is another simulated face mask dataset, based on images from FFHQ~\cite{ffhq}, created through a mask-to-face deformable model, which exploits multiple face landmarks to position a mask patch in order to simulate different wearing scenarios (correctly worn and three progressive incorrect wearing states).
Major problems affecting this kind of approach are the positioning of the masks on faces not facing frontally and the variety of masks in shape and texture.
We solved these issues by proposing an algorithm, based on Blender~\cite{blender} and MakeHuman~\cite{makehuman}, able to generate a completely synthetic dataset, in which it would be possible to modify at will the orientation of the head, the background, the size of the image and the shape and color of the masks applied to the body. 

\section{The Dataset}
In this section, we show how the \ours dataset has been created and which criteria we adopted to filter the datasets that constitute it.

\subsection{Problem Statement}
We focused on the face mask classification task on low-resolution images, ascribable to a classic video surveillance scenario. We have therefore selected $64\times64$ pixel as the maximum size of the faces, and used this to filter out images from all the selected sources.
This choice was made with a trade-off between a sufficiently small image size and the number of real images publicly available, besides being a common resolution for analysis on small images, see for instance the Tiny Imagenet Challenge \cite{tiny_imagenet}.
As reported in Sec.~\ref{sec:related} and Tab.~\ref{tab:datasets}, the number of small bounding box faces is limited in every dataset. \ours collect them all, realizing the largest low-resolution face mask dataset to date. This ensemble forms the \ours training set, which will be explained in the following section. In~Sec.~\ref{sec:ice} we will detail the \ours testing partition, which has been designed to cover as much as possible the whole intra-class variance of the face mask classification problem.  The analysis of the visual domain captured by these partitions will be shown in Sec.~\ref{sec:visual_domain}.

\begin{table}[!t]
\centering
\caption{List of datasets considered in the creation of \ours. Faces are considered \emph{Small} if $max(width, height)\leq64$.}
\label{tab:datasets}
\begin{tabular}{lrrrr}
\toprule 
\multicolumn{1}{l}{\textbf{Dataset}} & \textbf{Mask} & \textbf{No-Mask} & \textbf{Wrong-Mask} & \textbf{Small}  \\ 
\midrule
RMFRD~\cite{rwfmd}              & 6.664 & 0 & 0 & 1905            \\ 
MMD~\cite{mmd}                 & 2,855 & 581 & 119 & 986       \\ 
FMD~\cite{kagglefmd}           & 768 & 286 & 97 & 532          \\ 
Medical Mask~\cite{kagglemmd}  & 3,918 & 1,319 & 0 & 305       \\ 
FMLD~\cite{fmld}               & 29,561 & 43,840 & 1,531 & 11,856    \\ 
Moxa3K~\cite{moxa}             & 5,380 & 1,477 & 0 & 916           \\ 
\midrule
Ours (no synth.)               & 9,055 & 12,620 & 1701 & All   \\ 
\textbf{Ours}                  & 21,384 & 15,772 & 1701 & All  \\
\bottomrule
\end{tabular}
\end{table}

\subsection{The \ours Training Partition}
We collected the images from different datasets, chosen for their peculiarity such as the total number of images (FMLD), challenging environments (Moxa3K), specific types of face masks (Medical Mask, FMD, MMD), and the high number of native low-resolution images w.r.t. the total amount of that specific dataset (MMD, Moxa3K, RMFRD). 
The cardinality of each dataset is presented in Tab.~\ref{tab:datasets}. 
All the considered datasets are freely available to academia, as stated by their respective licenses. Other very interesting datasets for our task (\eg~\mbox{MAFA-FMD}~\cite{retinafacemask}), although present in literature, are not publicly available or under commercial licences.

We merged the above datasets' training sets into a single set, splitting it into two categories: \emph{Mask} for the faces correctly wearing a face mask, and \emph{No-Mask} otherwise.
The resulting size of the merged dataset is:
\begin{itemize}
    \item \emph{Mask}: 49,227 images
    \item \emph{No-Mask}: 47,503 images
\end{itemize}

We removed possible duplicates by analyzing all the images using the structural similarity index measure (SSIM~\cite{ssim}) and then we kept only the images whose size is lesser or equal to $64\times64$ px. As is, the resulting dataset has the following composition:
\begin{itemize}
    \item \emph{Mask}: 9,055 images
    \item \emph{No-Mask}: 12,620 images
\end{itemize}

\begin{figure}
\centering
    \includegraphics[width=0.7\linewidth]{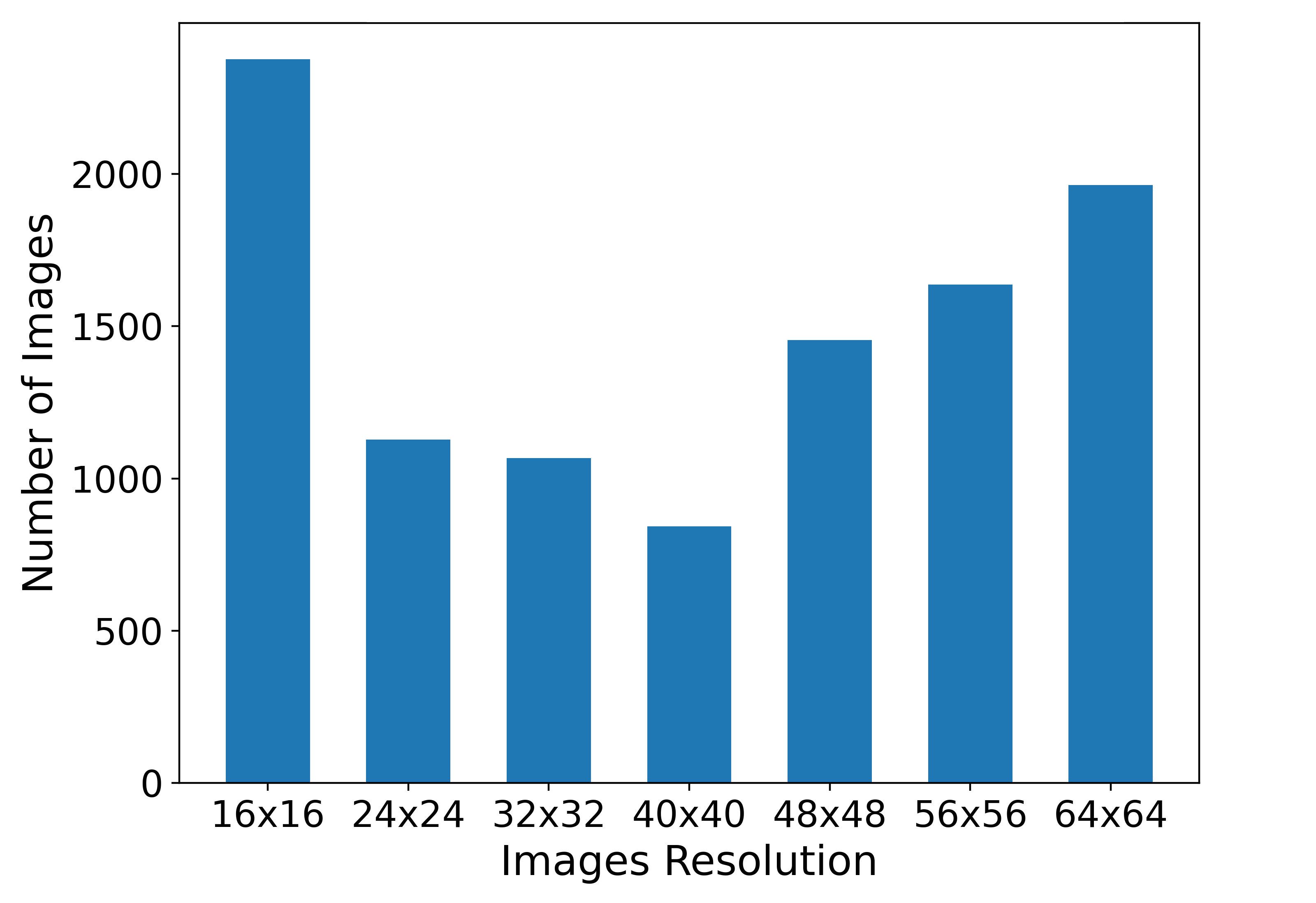}
    \caption{Distribution of image sizes in \ours (without synthetic data). Each bar represents images of resolution less or equal to that specific tick. The synthetic images are all of the same size and can be produced in any amount.}
    \label{fig:dataset_sizes_distribution}
\end{figure}

The distribution of \ours image sizes is reported in Fig.~\ref{fig:dataset_sizes_distribution}. Some dataset also gives the annotation of \emph{Wrong-Mask}, which is the class for face mask incorrectly worn, such as the mask below the nose, or under the chin. This class consists of 1,701 elements.

\subsection{Data Processing}
Some of the datasets used to compose \ours are already cropped on the full face, showing also the head of the person and not only the eyes, nose, and mouth. Some of the datasets that use bounding box notation, however, do not. FMD, FMLD, MMD, and Moxa3K provide bounding box annotations enabling face extraction. 
Nonetheless, some of the bounding boxes are tight around the face. In order to obtain the full face of the subject, we enlarged the bounding box on the full face, extending the width and height of the original bounding box by $\frac{1}{6}$ of each respective dimension (width, height).

The dataset RMFRD has been generated by searching the web for images, hence it required manual annotation. In particular, we first ran a face detector~\cite{zhang2016joint}, then we manually proceeded by removing the false positive detections. During the annotation process, the resulting bounding boxes extension has been performed as well.

\subsection{The \ours Testing Partition}\label{sec:ice}
In order to test the reliability of the dataset, an additional set of 1077 face images (584 \emph{Mask}, 270 \emph{No-Mask}, and 223 \emph{Wrong-Mask}) has been created from a video sequence acquired with multiple surveillance cameras in the ICE lab\footnote{\url{https://www.icelab.di.univr.it/}.} of the University of Verona. This dataset is our sample for the video surveillance scenario, and we used it as test set to inspect the actual accuracy of the classifiers trained over \ours. The dataset has been carefully acquired, and manually annotated to have a true example of video surveillance in a real case scenario. 
A few samples of this partition are presented in Fig.\ref{fig:dataset_composition}.

\begin{figure*}[h!]
    \centering
    \includegraphics[width=\textwidth]{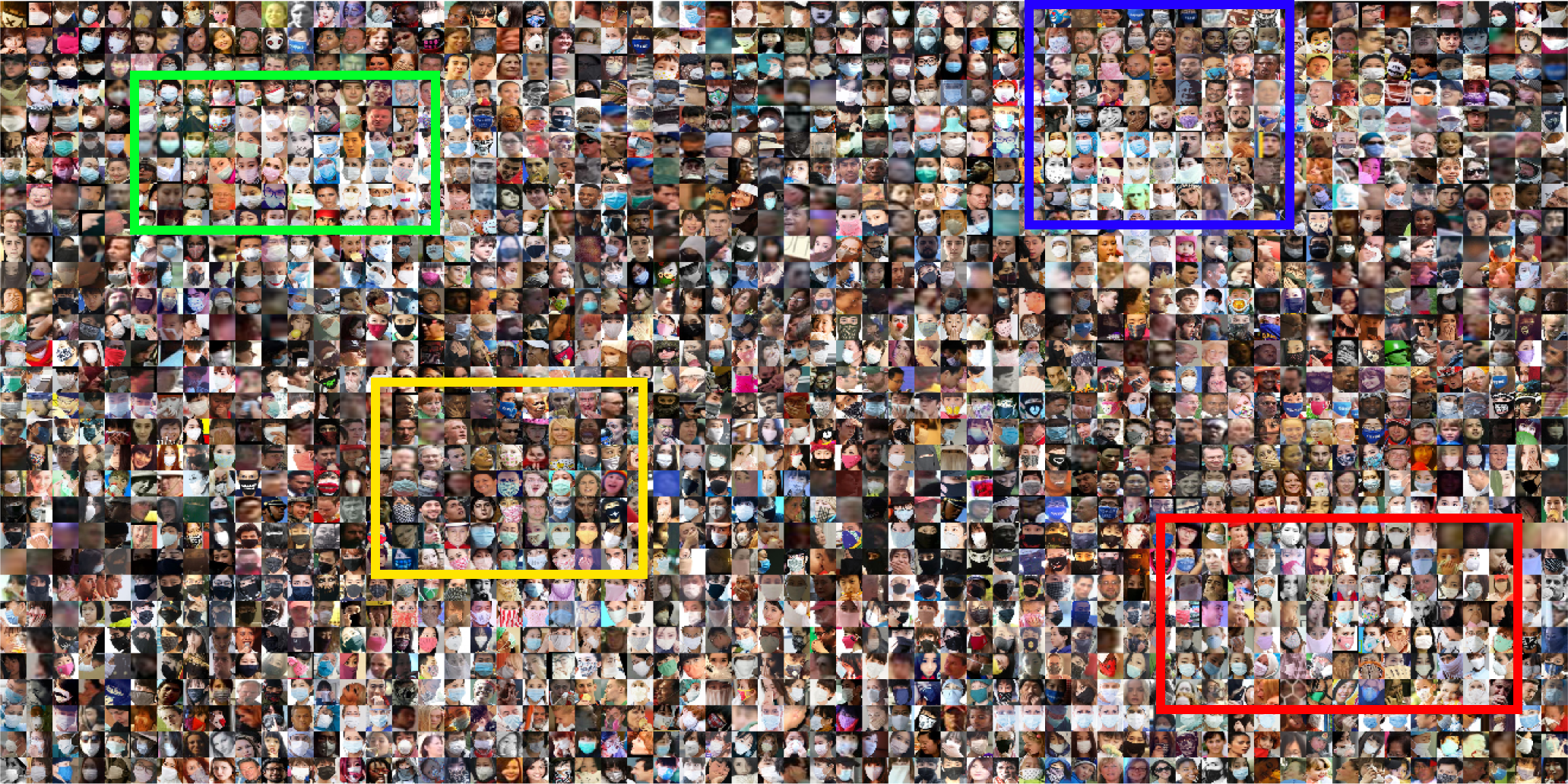}\label{subfig:cg}
    \caption{The CG approach on the \ours dataset. Here are highlighted different cluster areas in which there are neighborhood similarities. In particular: the top-left green area creates a cluster with a lot of faces facing to the left side, the top-right blue area has many straight-forward looking faces, the bottom right red area contains right-facing faces, and finally, the yellow area contains faces with an angle from below or similar. The full-resolution image will be presented on the dataset web page.}
    \label{fig:first_step}
\end{figure*}

\subsection{Visual Domain Analysis}\label{sec:visual_domain}
Not all the datasets contained in \ours present annotations such as head pose, gender, ethnicity or visual attributes of the masks (color, patterned etc.). For this purpose, we used an unsupervised method for manifold learning and human information interaction called Counting Grids (CG) \cite{perina2011image,lovato2013we}. The CG assumes the images are represented by an histogram of counts, and provides a torus (usually flattened as a rectangle with wrap-around) where images which are close indicate a smooth transition of some of their histogram's bins. The generative model underlying a CG is not straightforward, and we suggest reading~\cite{perina2011image} for a strict  mathematical analysis.

In our case, we extracted from each image different local descriptions, from quantized multi-level color SIFT descriptors~\cite{lowe1999object} to latent code of variational autoencoders~\cite{ramesh2021zero} without discovering big differences. In this paper we report the results using 300-dim SIFT codes. Then, we set up a $30\times60$ grid with an overlapping window of 11, essentially indicating the amount of overlap (in bins) where smoothness between adjacent grid locations is required. 

In Fig.~\ref{fig:first_step} we report the grid, where we can notice some clear patterns emerging from the dataset, with clusters of similar images and some smooth transitions from one area to another. Specifically, there are visible clusters of masked faces with different illumination conditions, native image resolution, and even similar face orientations. After a careful analysis of the grid (considering also alternative versions obtained with different parameterizations) we discovered very few images representing the typical surveillance scenario where people are observed slightly from above, with angled perspectives. To overcome this limitation we decided to extend our masked face dataset with a novel and dedicated synthetic dataset, presented in the next section.

\begin{figure}[!h]
\centering
    \subfloat[][]
    {\includegraphics[width=0.5\textwidth]{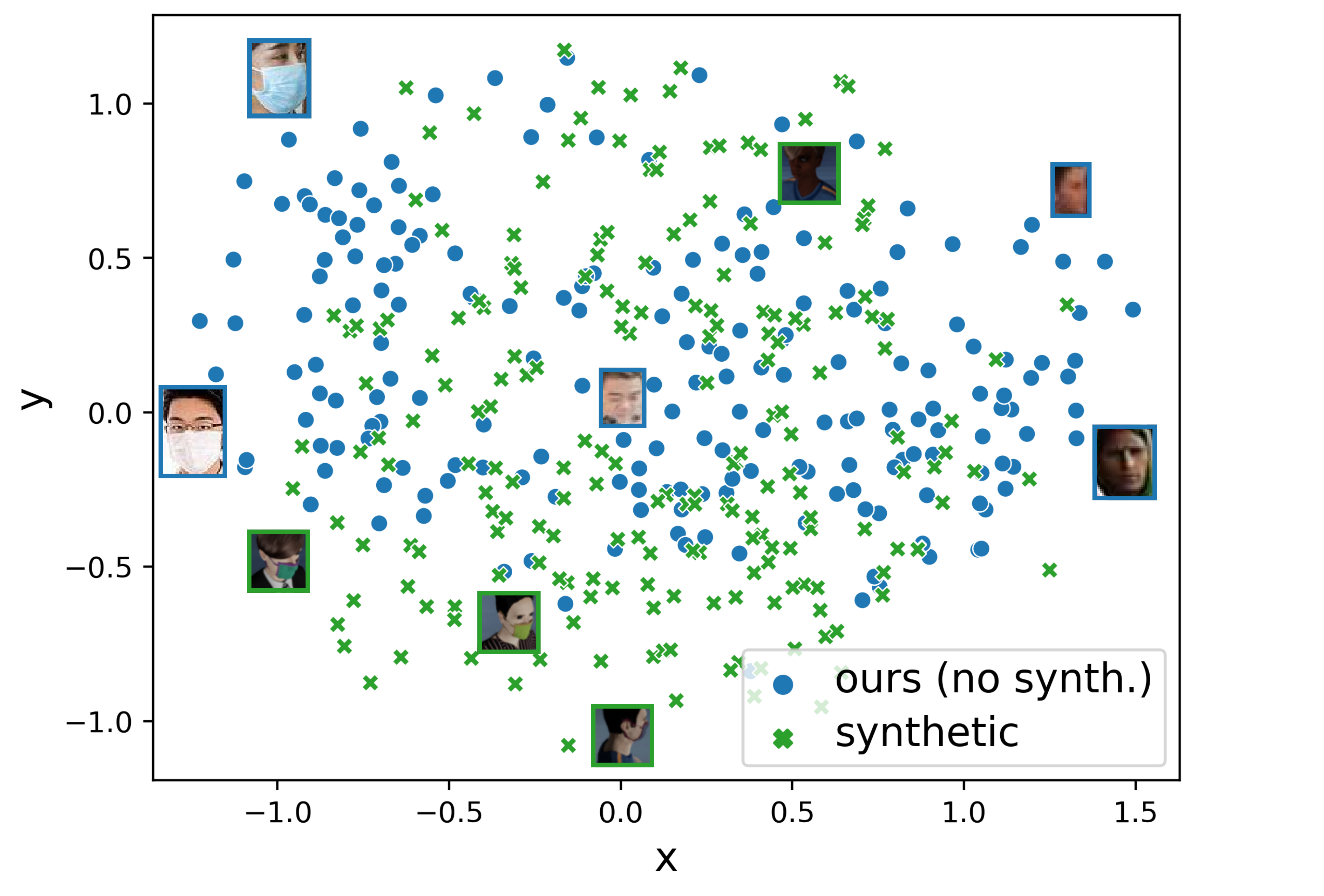}
    \label{subfig:pca_no_test}}
    \subfloat[][]
    {\includegraphics[width=0.5\textwidth]{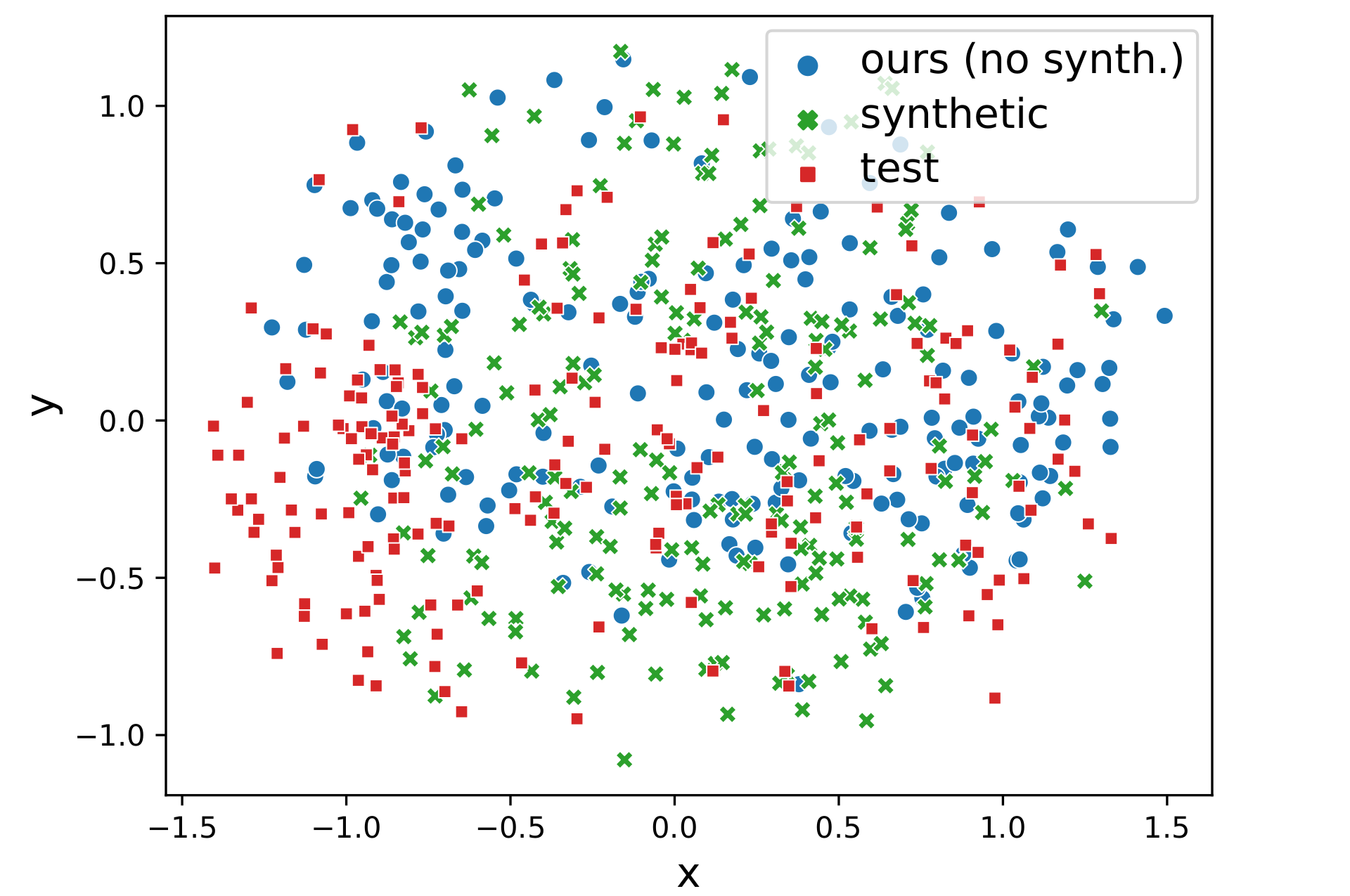}
    \label{subfig:pca_with_test}}
    \caption{Results of the Kernel PCA algorithm, comparing the \ours training partition, testing partition and the synthetic partition generated as in Sec.~\ref{sec:synth}. Fig~\ref{subfig:pca_no_test} shows that the generated data goes to fill an empty portion of the space (at the south). This enforces the idea that the generated fully synthetic data is capable to add variability to the dataset without changing its nature. In figure~\ref{subfig:pca_with_test}, it can be seen that the test set is actually ranging all over the variance of the training data.}
    \label{fig:kernelpca}
\end{figure}

\subsection{Synthetic Data Generation}\label{sec:synth}
To compensate for the variability in terms of camera viewpoints, as well as people and face masks appearance, we generated a fully synthetic dataset. Artificial human body models, textures, and backgrounds are therefore manageable via a parametric model. It is worth noting that the use of synthetic datasets to train models that cope with real test data has been already investigated, for example in re-identification~\cite{barbosa2018looking}. Other approaches have tried to extend real masked face datasets \cite{maskedfacenet,rwfmd} with custom-created data. Unfortunately, they usually leverage images extracted from large face recognition datasets and automatically apply mask patches on them, meaning that the user can only modify the appearance of the masks, without any control over the characteristics of the faces.
We instead developed a generation pipeline based on MakeHuman~\cite{makehuman}, a tool designed for the prototyping of photo-realistic humanoid models. With this tool and Blender~\cite{blender}, we created over 12000 synthetic human bodies, with different appearances in terms of age, ethnicity, gender, and clothing, and an image size coherent with \ours. We set the generation parameters to balance the gender and ethnicity distributions during the generation.

Starting from these synthetic people, we applied a face mask to 80\% of the total, setting them as \emph{Mask} samples, while the remaining 20\% is considered \emph{No-Mask}. We collected several texture patterns and randomly apply them to the face masks, in order to obtain variability in terms of face masks colors. Also, random backgrounds collected from the web have been used.
The typology of masks is another of the many customisable parameters: from FFP1 (surgical mask) to FFP2 and FFP3. For the first version of this synthetic dataset, we mostly used the FFP1 face mask type. 
Finally, we acquired one up to three different views of each synthetic human, angled from above, to fill the gap we have discovered with the counting grid. For this task we used Blender, in order to have full control over the image generation (such as illumination conditions, occlusions, etc).

\begin{figure}
    \centering
    \includegraphics[width=0.8\linewidth]{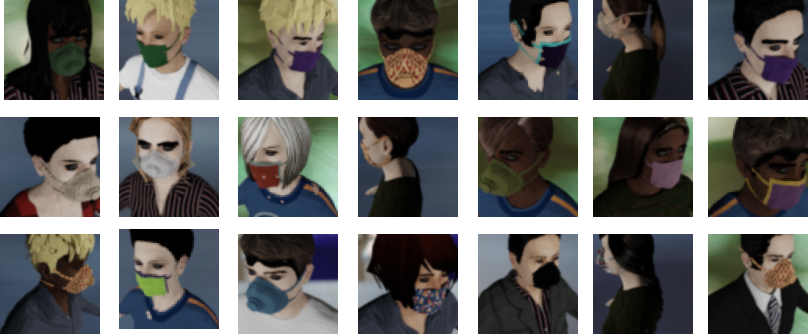}
    \caption{Example of some of the images in the \ours synthetic partition. It is possible to notice the variability in ethnicity of the people as well as the different colors, patterns and types of face masks applied on the models.}
    \label{fig:synthetic_example}
\end{figure}

Examples of synthetic image generation are shown in Fig.~\ref{fig:synthetic_example}. The generation scripts will be released along with the dataset at the same link.
To assess the usefulness of a synthetic dataset created following these guidelines, we decided to perform an analysis of the distribution of the data. Initially, all the images were resized to a $64\times64$ px resolution and feature maps were extracted with the same ResNet-50~\cite{resnet} network trained for the experiments. A random sample of the data divided by subset (\ours training partition, synthetic data and \ours test partition) was used. We then applied Kernel PCA~\cite{kpca} with a linear kernel. As visible in Fig.~\ref{subfig:pca_with_test}, the two subsets, the \ours training partition and the synthetic one, both contribute to filling different areas of the two-dimensional space. The conjunction of the two then results in a valid representation of the image space, hence providing a suitable training dataset for the classification task. Moreover, from Fig.~\ref{subfig:pca_with_test} it is evident that the \ours test partition actually covers almost all the feature space, validating the quality and the variability of the test images.

\section{Experiments}
Two main experiments have been carried out. In the first one, we validate the expressivity of \ours with and without synthetic data by training several classifiers, and testing on our test set. As a comparison, we trained the same classifiers on each of the datasets that compose \ours. 

We used several architectures including VGG19~\cite{vgg19}, Resnet-50~\cite{resnet}, MobileNetv2~\cite{mobilenetv2}, and EfficientNet~\cite{efficientnet}. The results in Table \ref{tab:training} show that the classifiers trained on \mbox{\ours} without synthetic data are often performing better, as in the case of MobileNet and EfficientNet. The introduction of the synthetic data gives a general boost, validating the manifold learning analysis and the successive synthetic data generation, which has focused on specific poses, that were missing in the original collection of datasets.

\begin{table}[]
\centering
\caption{Results obtained with different classifiers, trained on the different datasets, and tested with our benchmark acquired in the ICE lab (see Sec.~\ref{sec:ice}). Since RMFRD does not have any \emph{No-Mask} images, it was not possible to perform a comparable test.}
\label{tab:training}

\begin{tabular}{lrrrr}
\toprule
\multicolumn{1}{l}{\textbf{Dataset}} & \multicolumn{1}{c}{\textbf{ResNet-50}} & \multicolumn{1}{c}{\textbf{VGG19}} & \multicolumn{1}{c}{\textbf{MobileNet}} & \multicolumn{1}{c}{\textbf{EfficientNet}} \\
\midrule
MMD~\cite{mmd}                 & 0.743 & 0.848 & 0.822 & 0.778 \\
FMD~\cite{kagglefmd}           & 0.809 & 0.820 & 0.834 & 0.837 \\
Medical Mask~\cite{kagglemmd}  & 0.837 & 0.840 & 0.788 & 0.849 \\
FMLD~\cite{fmld}               & 0.842 & 0.856 & 0.772 & 0.857 \\
Moxa3K~\cite{moxa}             & 0.849 & 0.835 & 0.846 & 0.839 \\ 
\midrule
Ours (no synth.)                & 0.806 & 0.845 & 0.855 & 0.879 \\ 
\textbf{Ours}                  & \textbf{0.864} & \textbf{0.873} & \textbf{0.857} & \textbf{0.884} \\
\bottomrule
\end{tabular}
\end{table}

In the second experiment, we performed a leave-one-out pipeline, where we trained with all but one dataset, which was used afterward as an additional testing set, besides the \ours testing partition. 

Results are visible in Table~\ref{tab:leave_one_out}, showing that the most effective dataset among the pool composing the \ours training partition is indeed the FMLD dataset. This is probably due to the fact that it is the largest dataset.
These experiments have been conducted on a ResNet-50 architecture.

\begin{table}[]
\centering
\caption{Leave-one-out experiments on the datasets. The results show that FMLD is the decisive dataset for the leave-one-out strategy, probably due to the large number of images contained in it. As in Table~\ref{tab:training}, no results on RMFRD are reported here, as it does not contain any image of class \emph{No-Mask}.}
\label{tab:leave_one_out}
\begin{tabular}{lrr}
\toprule
\textbf{Dataset Left Out} & \textbf{Left Out Acc.} & \textbf{Test Set Acc.} \\ \midrule
MMD~\cite{mmd}                & 0.982 & 0.823 \\
FMD~\cite{kagglefmd}          & 0.961 & 0.822 \\
Medical Mask~\cite{kagglemmd} & 0.960 & 0.827 \\
FMLD~\cite{fmld}              & 0.812 & 0.801 \\
Moxa3K~\cite{moxa}            & 0.928 & 0.835 \\
\bottomrule
\end{tabular}
\end{table}

\section{Conclusions}
In this paper, we present \ours, a novel dataset of native low-resolution images of masked faces, divided into three classes indicating the presence, absence, or incorrect placement of the face masks. We also provide a synthetic data generation method to balance the dataset in terms of the number of images per class and attributes such as gender, ethnicity, face mask variability, illumination conditions, and camera point-of-view. This dataset has proven to be a valid benchmark for surveillance scenarios, in which the people occupy a small portion of the image, and the faces themselves occupy even fewer pixels.

\subsubsection*{Acknowledgements}
This work was partially supported by the Italian MIUR within PRIN 2017, Project Grant 20172BH297: I-MALL - improving the customer experience in stores by intelligent computer vision and the POR FESR 2014-2020 Work Program of the Veneto Region (Action 1.1.4) through the project No. 10288513 titled ``SAFE PLACE. Sistemi IoT per ambienti di vita salubri e sicuri''.

\bibliographystyle{splncs04}
\bibliography{bibi}

\end{document}